\documentclass[12pt]{article}

\usepackage{fullpage}
\usepackage[hmargin=1.0in,vmargin=1.0in]{geometry}

\usepackage{graphicx}
\usepackage{float}
\usepackage{amsmath}
\usepackage{amssymb}
\usepackage{amsthm}
\usepackage{latexsym}
\usepackage{verbatim}
\usepackage{threeparttable}
\usepackage{multirow}
\usepackage{caption}
\usepackage{booktabs}

\usepackage{authblk}
\usepackage{fancyhdr}
\usepackage{natbib}
\usepackage{url}
\usepackage{hyperref}

\usepackage{listings}
\usepackage{color}
\usepackage[authormarkup=none,commandnameprefix=ifneeded]{changes}

\usepackage{todonotes}

\usepackage{titlesec}
\titleformat*{\section}{\LARGE\bfseries}
\titleformat*{\subsection}{\Large\bfseries}
\titleformat*{\subsubsection}{\large\bfseries}

\usepackage{lineno}
\usepackage{setspace}

\begin{document}
% \linenumbers
\doublespacing

%\listofchanges[style=<list|summary>]
% \title{An integration of pig growth and whole-genome prediction models by leveraging Bayesian hierarchical models}
\title{Evaluating transfer learning strategies for improving dairy cattle body weight prediction in small farms using depth-image and point-cloud data}

\author[1]{Jin Wang}
\author[1]{Angelo De Castro}
\author[1]{Yuxi Zhang}
\author[1]{Lucas Basolli Borsatto}
\author[1]{Yuechen Guo}
\author[2]{Victoria Bastos Primo}
\author[2]{Ana Beatriz Montevecchio Bernardino}
\author[3]{Gota Morota}
\author[2]{Ricardo C Chebel}
\author[1*]{Haipeng Yu}
\affil[1]{Department of Animal Sciences, University of Florida, Gainesville, FL, USA 32611}
\affil[2]{Department of Large Animal Clinical Sciences, University of Florida, Gainesville, FL, 32611}
\affil[3]{Laboratory of Biometry and Bioinformatics, Department of Agricultural and Environmental Biology, Graduate School of Agricultural and Life Sciences, The University of Tokyo, Bunkyo, Tokyo 113-8657, Japan}

\date{}

\maketitle

\newpage 

\noindent
% Running title:  \\

\noindent 
* Corresponding author: \\
Haipeng Yu\\
Department of Animal Sciences \\
University of Florida \\
Gainesville, Florida 32611 USA. \\
E-mail: haipengyu@ufl.edu\\

\newpage
\section*{Abstract}
Computer vision–based approaches have emerged as automated, non-invasive, and scalable tools for monitoring dairy cattle, thereby supporting effective management, health assessment, and phenotypic data collection. Although transfer learning is commonly employed in studies predicting body weight from images, its effectiveness and optimal fine-tuning strategies remain poorly understood in livestock applications, particularly beyond the straightforward use of pretrained ImageNet or COCO weights. In addition, while both depth images and three-dimensional point-cloud data have been explored for body weight prediction, direct, head-to-head comparisons of these two modalities in dairy cattle are limited. Therefore, the objectives of this study were to 1) evaluate whether transfer learning from a large farm can enhance body weight prediction performance on a small farm with limited data, and 2) compare the predictive performance of depth-image- and point-cloud-based approaches under three experimental designs. Top-view depth images and point-cloud data were collected from 1,201, 215, and 58 cows at large, medium, and small dairy farms, respectively. Four deep learning models were evaluated: ConvNeXt and MobileViT for depth images, and PointNet and DGCNN for point clouds. We found that transfer learning markedly improved body weight prediction on the small-farm test set across all four models, outperforming single-source learning and achieving gains comparable to, or greater than, those obtained through joint learning. These results indicate that pretrained representations can generalize effectively across farms with differing imaging conditions and dairy cattle populations. No consistent performance difference was observed between depth-image– and point-cloud–based models. Taken together, our findings suggest that transfer learning is a particularly suitable strategy for small-farm prediction scenarios where cross-farm data sharing is constrained by privacy, logistical, or policy considerations, as it requires access only to pretrained model weights rather than raw data.

\newpage
\section*{Introduction}
Regular monitoring of body weight (BW) in dairy cattle is essential for herd management, health assessment, and productivity \citep{van2010time}. Accurate BW information supports early detection of health issues, optimization of feeding strategies, and informed breeding decisions. Traditional BW measurement methods, such as walk-over scales or static weighing systems, are labor-intensive, time-consuming, and often impractical for frequent use on commercial farms. To address these limitations, precision livestock farming technologies have emerged, providing automated and non-invasive solutions for animal monitoring \citep{poursaberi2010real,yu2021forecasting, wang2024shinyanimalcv, de2025animalmotionviz}. Among these technologies, computer vision (CV) has gained increasing attention as a scalable approach for predicting cattle BW by analyzing images or video data \citep{miller2019using, bi2023depth}. CV-based BW prediction eliminates the need for physical contact, thereby reducing labor requirements while enabling more frequent and consistent data collection \citep{qiao2021intelligent}.

Transfer learning is often used in CV analysis, where a deep learning model trained on one task or dataset is reused and fine-tuned for a different but related task \citep{yosinski2014transferable}. It leverages previously learned features to improve performance and reduce data requirements in the new task. In livestock CV settings, the ImageNet \citep{deng2009imagenet} and COCO \citep{lin2014microsoft} datasets are common sources of pretrained convolutional neural network backbones \citep{rodriguez2019estimating,espejo2020towards,machuve2022poultry,ruchay2024barn, bi2025industry}. However, the potential of transfer learning is not limited to the use of pretrained weights from ImageNet and COCO. For example, deploying CV-based BW prediction models requires large-scale labeled datasets, which presents challenges for small farms with fewer animals and limited image data, often leading to suboptimal model performance when training models from scratch. In such cases, transfer learning can be used to leverage large farm datasets to enhance prediction on smaller target farm datasets. However, its effectiveness for improving BW prediction on small dairy farms remains underexplored. One challenge is determining how to fine-tune transfer learning effectively, given that CV phenotyping setups vary across farms, with differing levels of image quality and phenotyping noise.

Among the various CV approaches, depth images have been widely used for BW prediction because they can capture the height and contour information of animals. Many existing studies extract morphological features, such as body length, width, height, and surface area, from segmented depth images and use these features as inputs to regression models for predicting BW \citep{kashiha2014automatic,yu2021forecasting, bi2023depth, xie2024novel}. More recently, deep learning methods have become increasingly prevalent for BW prediction tasks \citep{gjergji2020deep, dohmen2021image, bi2025industry}. Instead of relying on manually engineered features, these models learn hierarchical representations directly from raw depth images. In parallel, three-dimensional (3D) point cloud data have gained attention for their richer spatial representation of animal structure. Several preliminary studies have demonstrated the feasibility of using point cloud–based deep learning models for predicting cattle body volume or BW. For example, \cite{hou2023body} applied a PointNet++ architecture to cattle point clouds and reported accurate BW predictions. Related work has also explored point cloud pipelines for predicting cattle BW and size in practical settings \citep{yang2022automated}. However, these studies do not include direct comparisons between point cloud- and depth image-based approaches, and the relative utility of point cloud data for BW prediction remains insufficiently explored.

To address these gaps, the objectives of this study were 1) to investigate potential strategies for fine-tuning transfer learning from a large-farm dataset to improve BW prediction performance for a small-farm, and 2) to compare the predictive performance of depth images and point cloud data for dairy cattle BW prediction. To accomplish these objectives, top-view depth images and point cloud data were collected from three dairy farms of varying sizes and phenotyping system setups. Predictive models were then developed and evaluated to assess the effects of data modality and transfer learning on BW prediction performance.

\newpage
\section*{Materials and Methods}

\subsection*{Data collection design}
All animal procedures conducted for data collection were approved by the University of Florida Institutional Animal Care and Use Committee (IACUC protocol number 202200000534). Data were collected from three dairy farms: a large and a medium commercial farm in central Georgia and a small-farm in Florida. Image acquisition took place on November 7–9, 2024, and February 7–8, 2025, at the large-farm; February 8–20, 2025, at the medium-farm; and July 14 and 17, 2023, at the small-farm.

A custom-designed image acquisition pipeline was deployed at each farm to capture top-view depth images of dairy cows as they passed through a chute at the exit of the milking parlor. The system combined RFID-based identification with depth imaging: an RFID reader detected each animal’s electronic ear tag and associated each captured image with its electronic identification. A top-mounted Intel RealSense D455 camera provided synchronized RGB and depth streams, and acquisition was triggered when the animal was fully positioned beneath the camera. The depth stream was saved as depth CSV files for each captured image. To obtain scale-based BW simultaneously with imaging, a calibrated in-chute electronic scale was installed below the depth camera. The scale was connected to a weight reader that mounted on the side of the chute. As each cow entered the imaging zone, the chute gate was briefly closed so that the animal remained on the scale until the indicator displayed a stable reading. 

Across the three farms, the large-farm contributed 1,201 cows and 13,357 images, the medium-farm 215 cows and 2,116 images, and the small-farm 58 cows and 211 images. Recorded BW ranged from 530–980 kg (ages 3–10 years) at the large-farm, 487–976 kg (ages 2–8 years) at the medium-farm, and 514–903 kg (ages 2–7 years) at the small-farm.

\subsection*{Point cloud generation from depth CSV file}
The depth CSV files were converted into 3D point clouds using the built-in calibration parameters of the Intel RealSense D455 camera installed in each farm. For each camera, the built-in calibration provides two key sets of parameters. First, the horizontal and vertical focal lengths in pixels $(f_x, f_y)$ describe how many pixels correspond to a given angular deviation from the optical axis in the horizontal and vertical directions, and therefore determine how image coordinates are related to viewing directions. Second, the principal point $(c_x, c_y)$ gives the pixel coordinates where the optical axis intersects the image plane. 

Each depth CSV file can be viewed as a depth image. Each pixel $(x', y')$ in this depth image has an associated depth value $D(x', y')$, defined as the distance from the camera to the object surface at that pixel, either the cow's dorsal surface or the floor, along the viewing direction. The height of the animal at pixel $(x', y')$ is given by $Z(x', y') = H_{\text{camera}} - D(x', y')$, where $H_{\text{camera}}$ is the distance between the camera and the ground and $D(x', y')$ is the distance between the camera and the cow’s dorsal surface.

The calibration parameters $(f_x, f_y, c_x, c_y)$ were then used to convert each pixel coordinate into a viewing direction. First, the pixel coordinates were centered and normalized as
\begin{align*}
x_n &= \frac{x' - c_x}{f_x}, \label{eq:xn}\\
y_n &= \frac{y' - c_y}{f_y}, 
\end{align*}
which describes the horizontal and vertical direction of the ray leaving the camera through pixel $(x', y')$. The three-dimensional point corresponding to pixel $(x', y')$ was then obtained by scaling this direction vector by the height $Z(x', y')$ of the animal at that pixel.
\begin{equation*}
\begin{bmatrix}
X(x', y') \\[2pt] Y(x', y') \\[2pt] Z(x', y')
\end{bmatrix}
=
Z(x', y')
\begin{bmatrix}
x_n \\[2pt] y_n \\[2pt] 1
\end{bmatrix},
\label{eq:pointcloud}
\end{equation*}
where $X$ and $Y$ denote the horizontal and vertical coordinates and $Z$ denotes the height of the point above the floor. The camera calibration parameters $(f_x, f_y, c_x, c_y)$ and the height $H_{\text{camera}}$ used for point cloud generation at each farm are summarized in Table \ref{camera_intrinsics}. The resulting point clouds provided a raw 3D representation of the dorsal surface of each cow and were subsequently processed for model training.

\subsection*{Data preprocessing}
Raw depth values were clipped to [0, 10,000] millimeters (mm) to suppress spurious outliers from noise and reflective surfaces. The clipped depths were then linearly rescaled to $[0,1]$ to standardize the dynamic range across farms and facilitate stable optimization. Depth matrices were converted to grayscale images, resized to $224 \times 224$ pixels, and duplicated across three channels to match the input requirements of networks pretrained on ImageNet, while preserving the structural information of the original depth data.

Point clouds reconstructed from depth CSV files were first cleaned by removing non-finite points. Each cloud was then centered by subtracting its centroid and normalized to a unit sphere by dividing by the maximum Euclidean distance from the origin. This translation–scale normalization is standard in point-cloud analysis as it removes trivial pose and scale variability, improves numerical conditioning, and reduces sensitivity to farm-specific camera installation heights and the distance from animal to the camera \citep{qi2017pointnet, qi2017pointnet++, wang2019dynamic}. For efficient batching and to reduce point cloud computational cost, every cloud was standardized to 1,024 points. We used a lightweight hybrid subsampling strategy: (i) a stride-like selection to obtain approximately uniform coverage of the cloud, followed by (ii) random completion to reach the target cardinality when needed. This combination provides broad spatial coverage at low overhead and aligns with common training configurations in which 1,024 points balance accuracy and throughput \citep{qi2017pointnet, qi2017pointnet++, wang2019dynamic}. As with the depth preprocessing, all point-cloud steps were executed after establishing cow-level partitions to prevent information leakage.

\subsection*{Body weight prediction deep learning models}
\subsubsection*{Depth CSV models}
Motivated by the proven transferability of ImageNet–pretrained vision backbones \citep{deng2009imagenet}, we evaluated two depth image models with transfer learning: ConvNeXt-Tiny \citep{liu2022convnet} as a modern convolutional network and MobileViT-Small as a lightweight convolution–transformer hybrid \citep{mehta2021mobilevit}. Both backbones were initialized with ImageNet weights and adapted from classification to scalar regression using a unified modification. The original classifier was removed, the backbone served as a feature extractor, and a regression head was appended comprising global average pooling, a fully connected layer with ReLU activation, dropout regularization, and a single linear unit to output BW. Inputs were $224 \times 224 \times 3$ tensors obtained by duplicating the single channel depth map across three channels.

ConvNeXt follows a ResNet-style architecture but incorporates several modern design elements, such as large-kernel depthwise convolutions, inverted bottlenecks, LayerNorm within convolutional blocks, and a patchified stem, to match transformer-era performance while retaining the efficiency of pure convolutional neural networks \citep{liu2022convnet}. We chose ConvNeXt as a strong convolutional backbone that can exploit local geometric structure in depth images while approaching transformer-level accuracy, providing a natural reference model for depth-based BW prediction. In particular, we used the Tiny configuration as a compact yet accurate backbone for depth-image regression. This model was chosen because an improved ConvNeXt variant has also been used for non-destructive grading of preserved eggs, where it achieved higher internal-quality classification accuracy than conventional convolutional neural network baselines \citep{tang2024non}, further supporting the suitability of ConvNeXt-style backbones for agriculture-related imaging tasks.

MobileViT is a hybrid vision backbone that couples MobileNet-style inverted residual blocks, providing strong local inductive bias and parameter efficiency, with lightweight transformer blocks that capture long-range dependencies via patchwise self-attention \citep{mehta2021mobilevit}. A MobileViT block first encodes local features with convolutions, unfolds the feature map into non-overlapping patches for transformer processing, and then refolds and fuses the output with the convolutional stream to preserve fine spatial detail while adding global context. We used the MobileViT-Small configuration as a compact backbone for depth image regression. MobileViT was selected because it has demonstrated superior performance for BW prediction in pigs \citep{bi2025industry}.

\subsubsection*{Point cloud models}
To leverage geometric information directly from 3D point clouds, we used two foundational point cloud architectures, PointNet \citep{qi2017pointnet} and Dynamic Graph CNN (DGCNN) \citep{wang2019dynamic}, and adapted them for BW regression with modifications to improve computational efficiency. Inputs were preprocessed by centering at the centroid and normalizing to a unit sphere, retaining only X, Y, and Z coordinates without normals or intensity channels.

PointNet pioneered direct learning on unordered point sets by using symmetric functions to achieve permutation invariance \citep{qi2017pointnet}. The architecture applies shared multi-layer perceptrons to transform each point independently, then aggregates features via max pooling to produce a global shape descriptor invariant to input ordering. To handle pose variations, PointNet incorporates spatial transformer networks, known as T-Nets. An input T-Net predicts a 3-by-3 affine transformation matrix to canonically align the raw point cloud, while an optional feature T-Net learns a 128 $\times$ 128 transformation in the intermediate feature space to normalize learned representations.  PointNet and its hierarchical variants have also been adopted in livestock phenotyping tasks, including point-cloud–based BW prediction in pigs and beef cattle, where they achieved high agreement with reference measurements. For example, \citet{paudel2023deep} reported $R^2$ around 0.94 for finishing-pig BW prediction with a PointNet-based regressor, while \citet{hou2023body} used a PointNet++ pipeline for beef cattle and obtained a mean absolute percentage error (MAPE) of about 3.2\%. These applications indicate that a PointNet-type architectures provide a well-established basis for 3D BW prediction in our study.

Our PointNet implementation includes the input T-Net by default and allows toggling the feature T-Net to balance representational capacity against computational cost, with this choice determined during hyperparameter optimization. The feature extraction pathway consists of shared one-dimensional convolutions that progressively expand channel dimensions from 3 input coordinates to $64$, $128$, $256$, and finally to the embedding dimension, which ranges from 256 to 1024 depending on the configuration. Each convolutional layer is followed by batch normalization and ReLU activations. Following global max pooling over the point dimension, the regression head processes the resulting global feature vector through a three-layer fully connected network. The first layer projects from the embedding dimension to $256$ channels with batch normalization, ReLU activation, and dropout. The second layer reduces to $128$ channels, again followed by batch normalization, ReLU activation, and dropout. The final layer outputs a single scalar value for BW prediction. Compared to the canonical PointNet, we adjust layer widths for compactness, add dropout for regularization, and replace the multi-class softmax classifier with this scalar regression output.

DGCNN introduces dynamic graph convolution to capture local geometric structure by constructing edge features in learned feature spaces \citep{wang2019dynamic}. At each layer, the network dynamically recomputes a $k$-nearest neighbor graph in the current feature space and aggregates information for each point from its neighbors using both their relative offsets and absolute coordinates, an operation commonly known as edge-based convolution. The relative displacements capture local shape variations, while absolute coordinates preserve global positional context. This dynamic neighborhood update allows the network to adapt its receptive field as features evolve through the layers, enabling effective learning of hierarchical geometric patterns. DGCNN has been applied to 3D plant point clouds, for example, to reconstruct missing leaf regions and classify plant structures, demonstrating that dynamic graph–based encoders can provide robust performance on biological point clouds with irregular geometry and occlusions \citep{ohamouddou2025ms}. These properties motivated our choice of DGCNN as a complementary backbone to PointNet for 3D BW prediction.

Our DGCNN implementation uses a lightweight backbone with three edge-based convolution layers instead of the original five-layer design. The three layers produce output channels of $32$, $64$, and $128$ respectively, using LeakyReLU activations for smoother gradients and between $15$ to $20$ neighbors per point for graph construction. Features from all edge-based convolution layers are concatenated and projected to an embedding dimension of either 256 or 512, determined during hyperparameter optimization. The global descriptor is formed by concatenating both max-pooled and average-pooled features across all points, rather than relying on max pooling alone, to capture richer statistical information about the point cloud geometry. The regression head then processes this concatenated global feature through a three-layer fully connected network. The first layer projects from twice the embedding dimension to $256$ channels with batch normalization, LeakyReLU activation, and dropout regularization. The second layer reduces to $64$ channels with batch normalization and LeakyReLU activation. The final layer outputs a single scalar value representing the predicted BW.

\subsection*{Experimental designs}
\begin{figure}[H]
    \centering  
    \includegraphics[width=\linewidth]{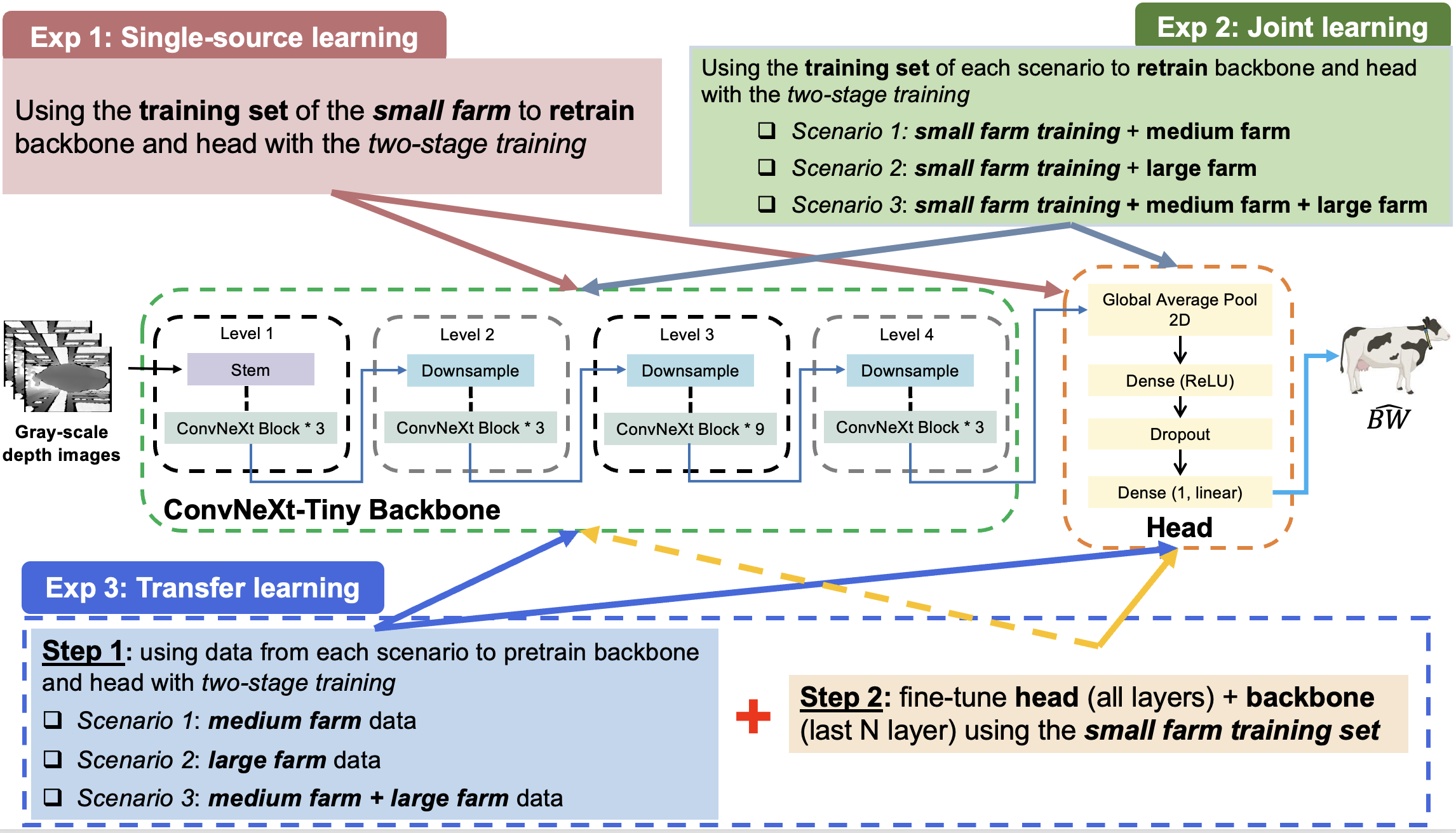}
    \caption{Three experimental designs were evaluated, together with their corresponding training-flow illustrations, using depth CSV files converted to images as model inputs. ConvNeXt-Tiny was used as an example backbone, followed by a shared regression head consisting of global average pooling, a fully connected layer with ReLU activation, dropout, and a final linear layer that outputs body weight. In the single-source learning setting, models were trained using only the small-farm training set. In joint learning, the small-farm training set was combined with data from the medium-farm, large-farm, or both to train all network layers end-to-end. Transfer learning was conducted in a two-step process. First, the backbone and regression head were pretrained using external training data. Second, the model was fine-tuned on the small-farm training set by updating the entire regression head and only the last N layers of the backbone. The value of N was determined through grid search using the small-farm validation set. Model performance was evaluated using predictions generated on the small-farm testing set.} 
    \label{experiment_design}
\end{figure}

\begin{figure}[H]
    \centering  
    \includegraphics[width=\linewidth]{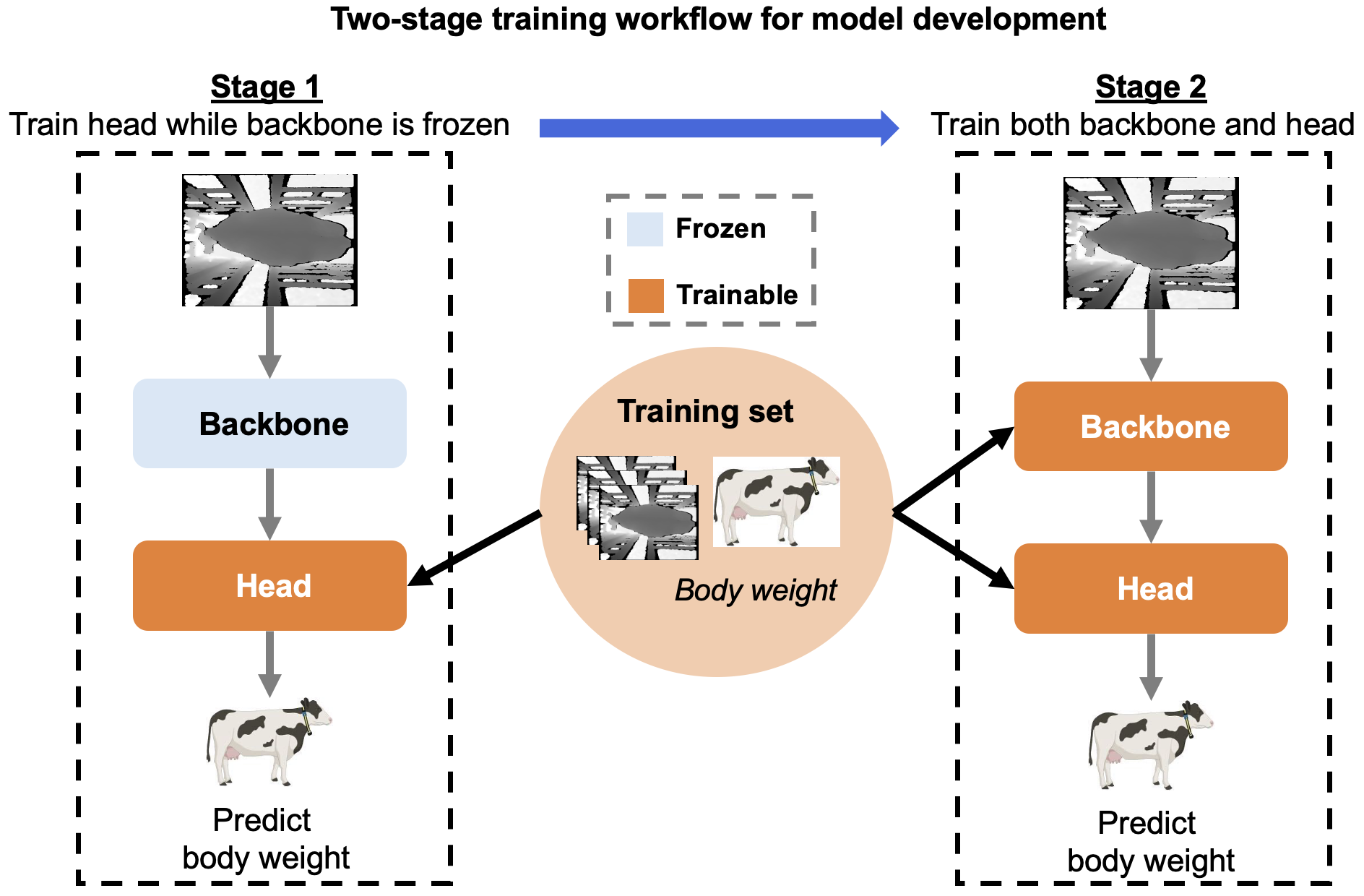}
    \caption{An illustration of the two-stage model training procedure. Stage 1 trains the model head while the backbone is frozen. Stage 2 trains both the backbone and the head together. This procedure was applied to the single-source learning design, the joint learning design, and the first step of the transfer learning design.} 
    \label{twostage}
\end{figure}

Our primary objective was to determine whether transfer learning improves prediction on the small-farm in Florida. We therefore evaluated three settings that progress from a within-farm baseline to designs that use external data from the large and medium farms in central Georgia. To highlight the benefit of transfer learning from the effect of simply adding more data, we also considered a joint-training approach by combining multiple data. Figure \ref{experiment_design} summarizes the three experimental settings and Figure \ref{twostage} shows the two-stage training protocol. All settings used five repeated random subsampling cross-validation. The goal of these three experimental designs was to examine whether different training strategies could improve prediction performance for the small-farm. All models trained under these three designs were evaluated using the same small-farm testing set, which remained fixed across all designs. In each run, cow-level train and test partitions were redrawn with an independent random seed and then held fixed across models and modalities, and results were aggregated over the five repeats.

\subsubsection*{Single-source learning design}
The whole small-farm data were first partitioned into training and testing sets at a 60:40 ratio. The training set (i.e., the full training set) was then further divided into subtraining and validation sets at an 80:20 ratio for hyperparameter tuning, with the search ranges summarized in Table \ref{hyperparameters_tuned}. The subtraining set was used to train the backbone and head, and predictions on the validation set were used to tune the hyperparameters and decide the final model structure (Figure \ref{experiment_design}). After the model structure was decided, the entire full training set was used to perform two-stage training (Figure \ref{twostage}) to stabilize optimization and support transfer \citep{gonccalves2025computer}. In stage one, the backbone was frozen and only the head was retrained. In stage two, the backbone was unfrozen and both the backbone and head were retrained together. After that, the model was tested using the small-farm testing set. The same procedure was also used for medium- and large-farm testing set prediction for comparison. Frame-level predictions from depth CSV files or point clouds were averaged per cow to obtain a single BW prediction.

\subsubsection*{Joint learning design}
To investigate whether gains in predicting the small-farm testing set can be attributed to increased training volume rather than transfer learning, we combined the small-farm training set with medium- and/or large-farm data. Three scenarios were considered for constructing the training sets. Scenario 1 included the small-farm training set plus all data from the medium-farm; Scenario 2 included the small-farm training set plus all data from the large-farm; and Scenario 3 included the small-farm training set plus all data from both the medium and large farms (Figure \ref{experiment_design}). Using Scenario 1 as an illustration, the procedure was similar to the single-source learning design. The key difference was that in single-source learning, the full training set came only from the small-farm, whereas in this joint-learning design, the full training set consisted of the small-farm training set plus all data from the medium-farm. During hyperparameter tuning, this full training set was split into subtraining and validation sets to tune hyperparameters and determine the model structure. This was followed by using the full training set to perform the same two-stage training described above (Figure \ref{twostage}). The final model was then tested on the same small-farm testing set.

\subsubsection*{Transfer learning design}
Similar to the joint learning design, three scenarios were considered for constructing the training sets (Figure \ref{experiment_design}). Scenario 1 included the small-farm training set plus all data from the medium-farm; Scenario 2 included the small-farm training set plus all data from the large-farm; and Scenario 3 included the small-farm training set plus all data from both the medium and large farms. Using Scenario 1 as an example, the full training set was the same as in the joint learning design, which was the small-farm training set plus all data from the medium-farm. Transfer learning was applied in two steps. In the first step, only the medium-farm data were used. These data were first split into subtraining and validation sets to tune hyperparameters and determine the model structure. Once the model structure was decided, the entire medium-farm dataset was used to perform the two-stage training described above (Figure \ref{twostage}). In the second step, the retrained model from the first step was further fine-tuned using the small-farm training data, where the whole head and the last few layers of the backbone were updated together. Specifically, for ConvNeXt and MobileViT, the number of unfrozen backbone layers was chosen from the last 10 to the last 100 in increments of five, while always training the regression head, and the choice was selected using validation MAPE. For the point cloud models, we compared head-only fine-tuning, head plus a small number of final backbone layers, and full-model fine-tuning, selecting the best option by validation MAPE. The optimal number of the last few backbone layers to be fine-tuned was selected using a grid search. Final evaluation in every cross-validation fold used the same small-farm training dataset as the other scenarios, with frame-level predictions averaged per cow.

\subsubsection*{Implementation}
All experiments were executed on the University of Florida HiperGator cluster using NVIDIA A100 and NVIDIA B200 GPUs, with mixed precision enabled when available. ConvNeXt was implemented in TensorFlow with Keras, MobileViT in PyTorch, and both point-cloud models in PyTorch. The depth image models used the Adam optimizer with Huber loss and the point-cloud models used AdamW with Smooth L1 loss and OneCycleLR during tuning. Early stopping and learning rate reduction on plateau were applied, and model selection used validation performance within each repeat.

\subsection*{Evaluation metrics}
Model performance was assessed on cows unseen during training. For single-source learning prediction, evaluation used each farm’s test cows. For joint learning and transfer learning, evaluation used the same small-farm test split as above to ensure comparability. Two metrics were reported: the coefficient of determination ($R^{2}$) and MAPE.

\begin{align*}
    & \textcolor{black}{R^{2} = 1 - \frac{\sum_{i=1}^{n} (y_i - \hat{y}_i)^2}{\sum_{i=1}^{n} (y_i - \bar{y})^2}} \\[0.3cm]
    & \textcolor{black}{\text{MAPE} = \frac{1}{n} \sum_{i=1}^{n} \left|\frac{y_i - \hat{y}_i}{y_i}\right| \times 100\% },
\end{align*}
where $y_i$ is the observed BW of cow $i$, $\hat{y}_i$ is the predicted BW of cow $i$, $\bar{y}$ is the mean of observed BW, and $n$ is the number of cows in the test set. $R^{2}$ reflects explained variance, and MAPE expresses the relative error as a percentage.

\newpage 
\section*{Results}
We report mean performance over five repeated subsampling cross-validations. Table~\ref{Overall_results} summarizes $R^2$ and MAPE for all models and scenarios. Because both metrics are scale-normalized, the results are comparable across farms with different BW ranges. 

\subsection*{Single-source prediction}
On the small-farm, DGCNN achieved the highest $R^2$ of 0.39 with a MAPE of 8.48\%, and PointNet achieved an $R^2$ of 0.37 with the lowest MAPE of 8.19\% (Table~\ref{Overall_results}). ConvNeXt achieved an $R^2$ of 0.30 with a MAPE of 9.32\%, and MobileViT-S achieved an $R^2$ of 0.16 with a MAPE of 10.02\%. Overall, the point cloud models performed better than the depth image models. For comparison purposes, on the medium-farm, MobileViT-S delivered the best performance, achieving an $R^2$ of 0.68 with a MAPE of 6.23\%. It was followed by DGCNN, ConvNeXt, and PointNet, in that order. No clear difference was observed between the depth image and point cloud models. On the large-farm, MobileViT-S was the best performer, achieving the highest $R^2$ of 0.85 with the lowest MAPE of 2.98\%. It was followed by DGCNN, PointNet, and ConvNeXt in terms of $R^2$, while DGCNN, ConvNeXt, and PointNet ranked in that order in terms of MAPE. Similar to the case of the medium-farm, no clear difference was observed between the depth image and point cloud models, although the point cloud models were less stable in terms of cross-validation uncertainty.

\subsection*{Joint learning with combined farms}
Compared with predicting the small-farm testing set under the single-source learning design, combined joint learning improved predictions for most models (Table~\ref{Overall_results}). When the small-farm training set was combined with the medium-farm, the best prediction was delivered by DGCNN, achieving an $R^2$ of 0.75, equivalent to a 92\% increase relative to the single-source learning design. ConvNeXt was the next-best model, followed by PointNet and MobileViT. The lowest MAPE was also achieved by DGCNN at 5.63\%, representing a 34\% reduction relative to the single-source learning design, followed by ConvNeXt, PointNet, and MobileViT. The ranking of the models was consistent between $R^2$ and MAPE.

When the small-farm training set was combined with the large-farm under the joint-learning framework, DGCNN again produced the best prediction, achieving an $R^2$ of 0.77, a 97\% increase relative to the single-source learning design. The second-best model was PointNet, followed by ConvNeXt and MobileViT. DGCNN also achieved the lowest MAPE at 5.38\%, equivalent to a 37\% reduction relative to the single-source learning design, followed by ConvNeXt, PointNet, and MobileViT.

When adding both the medium- and large-farm datasets to the small-farm training set, DGCNN delivered the highest $R^2$ of 0.74, corresponding to a 90\% increase relative to the single-source learning design, followed by PointNet, ConvNeXt, and MobileViT. For MAPE, the best-performing model was PointNet, followed by DGCNN, ConvNeXt, and MobileViT. PointNet achieved a MAPE of 5.74\%, representing a 30\% reduction relative to the single-source learning design.

Overall, combined joint learning delivered substantial prediction gains over the single-source learning scenario for predicting the small-farm testing set. In addition, the point cloud models consistently performed better than the depth image models when large-farm or combined medium- and large-farm data were included in the training set, whereas MobileViT was consistently the lowest performer, apparently having difficulty learning meaningful signal from heterogeneous data. DGCNN produced the best prediction performance in most cases.

\subsection*{Transfer learning}
Transfer learning improved performance beyond the baseline scenario, where no external data were added to the training set, and surpassed the combined joint learning design in some cases (Table~\ref{Overall_results}). Using the medium-farm as the source, DGCNN and PointNet tied with an $R^2$ of 0.70. However, their prediction $R^2$ values were slightly lower than those obtained under joint learning. The second-best model was MobileViT, followed by ConvNeXt. According to MAPE, the best model was DGCNN, with a MAPE of 5.73\%, which was slightly worse than joint learning. MobileViT was the next-best model, followed by PointNet and ConvNeXt. Although none of the best models from transfer learning exceeded the performance of those from joint learning in either $R^2$ or MAPE, transfer learning improved prediction performance relative to joint learning for MobileViT and PointNet, but not for ConvNeXt and DGCNN.

Adding the large-farm to the small-farm training data resulted in MobileViT being the best-performing model, with an $R^2$ of 0.78, which was slightly better than the best model from joint learning. The second best model was DGCNN, followed by PointNet and ConvNeXt. Based on MAPE, the best model was DGCNN with 5.07\%, representing a 6\% reduction in MAPE compared to the best performing model from joint learning. MobileViT, PointNet, and ConvNeXt followed next. Contrary to the medium-farm scenario, the best models from transfer learning were always better than those from joint learning. The benefit of transfer learning was more apparent: it improved the prediction performance of MobileViT, PointNet, and DGCNN relative to joint learning.

Lastly, MobileViT produced the best $R^2$ of 0.82 when adding the medium- and large-farm data to the small-farm training set. This was equivalent to an 11\% improvement relative to the best model in joint learning. The second-best model was DGCNN, followed by PointNet and ConvNeXt. Again, MobileViT was the best model according to MAPE, with a MAPE of 5.41\%, representing a 6\% reduction compared to the best model in joint learning. It was followed by DGCNN, ConvNeXt, and PointNet. Similar to the large-farm scenario, the best models from transfer learning were always better than those from joint learning. Except for PointNet, all models produced better $R^2$ and MAPE than their joint-learning counterparts.

Taken together, a consistent increase in $R^2$ and decrease in MAPE relative to the single-source design was observed for every model. Transfer learning consistently improved performance well beyond the small-farm baseline and, for MobileViT and in several DGCNN settings, matched or exceeded the corresponding gains from combined joint learning.

\newpage 
\section*{Discussion}
The success of CV-based precision livestock farming depends on ensuring that benefits reach not only large, well-resourced corporate farms but also smallholder farmers, who play a vital role in global food security and in sustaining the diversity and resilience of the agri-food system. Transfer learning is gaining momentum in deep learning, but its utility beyond simply using pretrained weights from ImageNet or COCO remains unclear in livestock CV studies. This study assessed whether transfer learning from larger farms can improve performance on a small farm using depth-image and point-cloud approaches for dairy cattle BW prediction across three experimental designs. 

The study also employed a cross-validation framework to evaluate the predictive performance of models on unseen data. Data partitioning for cross-validation can be implemented at either the image level or the animal level. In image-level cross-validation, images from the same animal may appear in both the training and testing sets, whereas animal-level partitioning does not allow the same animal to appear in both sets. All reported model predictive performance in this study was based on animal-level cross-validation to ensure that test animals were always unseen individuals, because image-level (i.e., non-animal-level) cross-validation is known to yield overly optimistic error estimates and to overestimate prediction $R^2$ \citep{wang2025impact}.

\subsection*{Impact of transfer learning strategies on BW prediction performance}
Relative to the single-source design, where the sole small-farm training set was used for model training, joint learning and transfer learning, that used external data improved prediction performance, although the magnitude and pattern of gains differed by approach and model. Under transfer learning, models were first pretrained on the external farm data and then fine-tuned on the small-farm training set with controlled layer unfreezing. Gains over the single-source design were substantial for all four models, and the extent of improvement depended on both the backbone and external data added. 

Overall, MobileViT benefited the most from transfer learning, as it did not perform well in joint learning. The rate of increase in $R^2$ from joint learning was 84\%, 160\%, and 165\%, respectively, when medium-, large-, and medium- and large-farm data were added to train models. This suggests that directly mixing different farm data in MobileViT was sensitive to how the small-farm split was sampled, whereas transfer learning yielded more stable and higher predictions. Similarly, the rate of reduction in MAPE from joint learning was 22\%, 43\%, and 59\%, respectively, when medium-, large-, and medium- and large-farm data were added. MobileViT was also often one of the best prediction models along with DGCNN. DGCNN was the most stable model, consistently performing well for both joint learning and transfer learning strategies. A mixed pattern was observed for ConvNeXt and PointNet, as they did not always show improvement under transfer learning.

Architectural differences between the four model backbones may provide an explanation for the stronger transfer gains observed for MobileViT and DGCNN on the small-farm. MobileViT is a lightweight hybrid backbone that couples MobileNet-style inverted residual blocks with transformer blocks operating on local patches, enabling it to reuse low-level convolutional filters while adapting long-range shape dependencies during fine-tuning \citep{mehta2021mobilevit}. This design has been shown to retain strong representational capacity with relatively few parameters, and recent applications of vision transformers in medical imaging report that pretraining followed by targeted fine-tuning can be particularly beneficial in small, domain-specific datasets, where attention layers can be repurposed without overfitting \citep{li2023transforming}. Similarly, DGCNN aggregates information over dynamically constructed $k$-nearest-neighbor graphs in feature space, so that each edge-based convolution layer learns relational features between neighboring points rather than treating points independently \citep{wang2019dynamic}. Studies on 3D point-cloud robustness have reported that such relation-based encoders retain performance better than PointNet-type architectures when point sampling, noise level, or local density vary between domains \citep{li2022improving}. In our setting, these properties are consistent with the larger transfer gains observed for MobileViT and DGCNN on the small-farm relative to ConvNeXt and PointNet, whose inductive biases are more strongly tied to either purely convolutional processing of 2D depth maps or point-wise processing of unordered point sets.

\begin{figure}[H]
    \centering  
    \includegraphics[width=\linewidth]{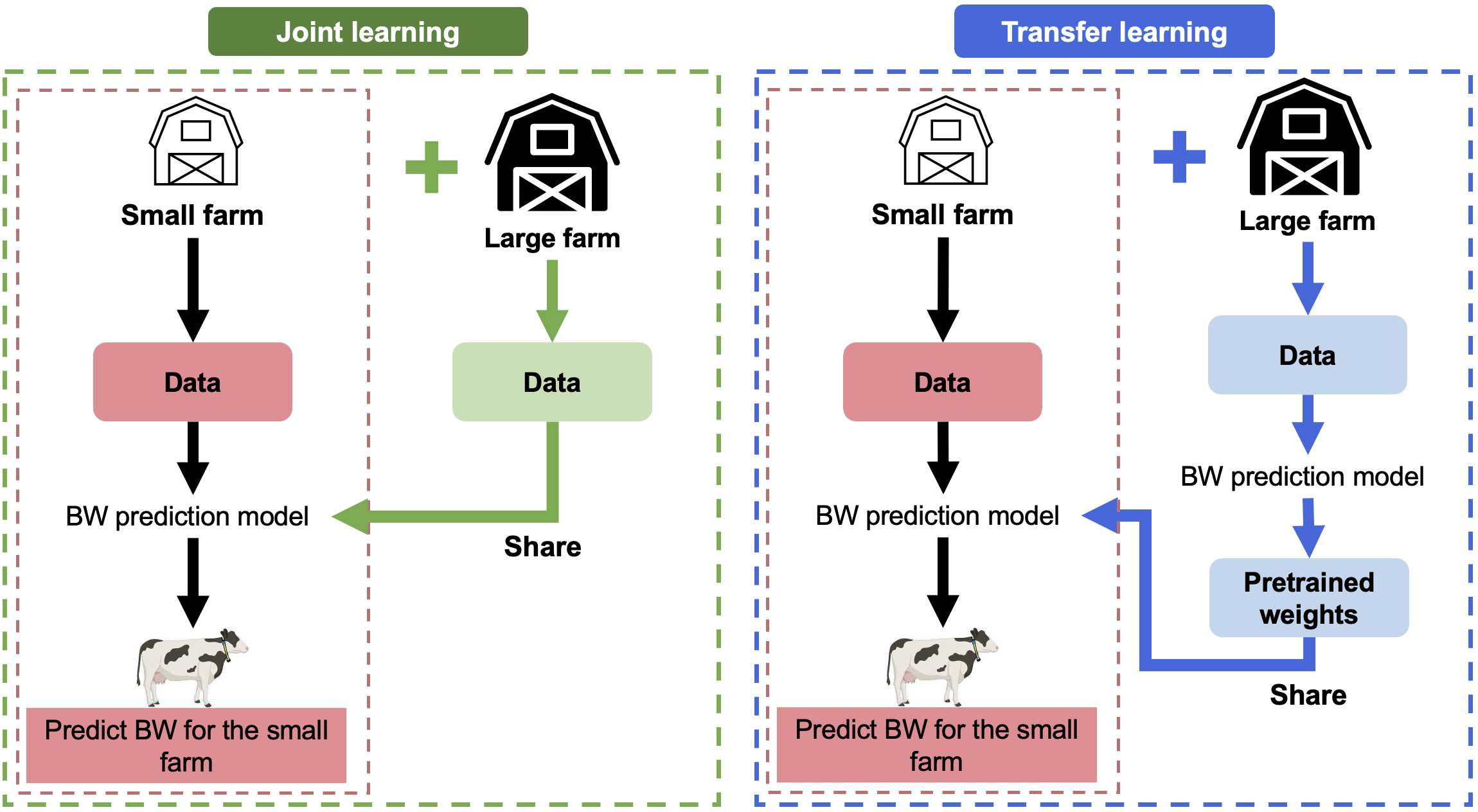}
    \caption{Comparison of joint learning and transfer learning for improving body weight (BW) prediction performance on the small-farm. Joint learning trains a BW prediction model by combining image data and scale-based body weights from both the small-farm and the external large-farm. This approach requires farms to share raw images and scale-based body weights. In contrast, transfer learning trains a model on the external farm and then provides only the pretrained model weights to the small-farm. The small-farm uses its own training data to complete model training. This approach avoids sharing raw data while still improving prediction performance for the small-farm.} 
    \label{jlvstl}
\end{figure}

Taken together, these patterns suggest that the benefit from external data is not solely a consequence of adding more samples. Simply mixing external farm data in joint learning can yield clear improvements relative to the single-source design, but pretraining followed by targeted fine-tuning can transfer meaningful biological signals more efficiently for some deep learning models in transfer learning and, in several cases, produced the best results for predicting the small-farm testing set without directly combining raw data. This distinction is practically important because joint learning requires different farms to share raw image data. However, in practice, many farms are unlikely to share images with others due to privacy, logistical, or policy constraints. The key advantage of transfer learning is that it only requires farms to share pretrained weights rather than raw data, while still conferring most of the attainable accuracy gains. The strength of transfer learning relative to join learning is illustrated in Figure \ref{jlvstl}. 

The observed advantage of transfer learning under limited target data aligns with prior CV studies in livestock that report marked improvements from deep transfer. For example, \citet{ruchay2024barn} fine-tuned ImageNet-initialized convolutional networks for on-barn cattle identification and reported identification accuracy above 94\% on barn-collected images, outperforming non-transferred baselines under the same conditions. \citet{machuve2022poultry} used transfer learning with modern convolutional backbones in poultry health tasks and achieved test accuracy of approximately 98\% in nondestructive fertility detection of chicken eggs, illustrating the strong gains achievable when pretraining is leveraged in small, domain-specific datasets. \cite{rodriguez2019estimating} predicted dairy cattle body condition scores from depth images and reported that transfer learning and light ensembling increased agreement with expert labels compared with training from scratch, reinforcing the value of transferred features when labeled target data are scarce.

In summary, our study found that when cross-farm data sharing is not possible, transfer learning is an effective substitute that often matches or exceeds the gains of joint learning with combined data, particularly for architectures whose features transfer well across farms, such as MobileViT and DGCNN. When combining data is permissible, enlarging the training set can yield additional reductions in error, such as ConvNeXt and PointNet, but the incremental benefit must be weighed against practical data sharing constraints.

\subsection*{Comparison of BW prediction performance between depth image and point cloud}
In the single-source design, the point cloud models performed better than the depth image models. This may suggest that when training data are limited, point cloud representations can capture 3D geometry more effectively than depth image models. However, this trend did not persist when switching to joint learning or transfer learning, where performance depended more on the model architecture itself. To our knowledge, no previous livestock studies have provided a head-to-head comparison of depth-image and point-cloud deep models for BW prediction on the same dataset. Instead, existing studies have typically adopted only a single modality. One factor that may explain the mixed performance of point cloud models is that our point clouds were standardized to 1,024 sampled points, a setting widely used in point cloud deep learning to balance representational fidelity and computational efficiency \citep{qi2017pointnet, qi2017pointnet++, wang2019dynamic} . Although this subsampling strategy preserves the global dorsal contour, it removes some local geometric detail that may be informative for BW prediction. In contrast, the depth image models receive dense two-dimensional projections that retain fine-scale contour and area cues, which may partly explain their stronger performance on the larger farms. An important direction for future research is to investigate whether varying the number of sampled points or using adaptive subsampling strategies can further improve point cloud model performance in livestock applications.

\newpage 
\section*{Conclusion}
This study investigated the effectiveness of transfer learning in improving dairy cattle BW prediction performance for small farms with limited data and compared the performance of depth images and point cloud data. We found that transfer learning substantially improved BW prediction on the small-farm testing set across all four models, with gains comparable to or exceeding those from combined joint learning. This suggests that pretrained representations can transfer effectively across farms with different imaging conditions and dairy cattle populations. We contend that transfer learning can be the preferred strategy for small farms when cross-farm data sharing is not feasible due to privacy, logistical, or policy constraints. Overall, either MobileViT or DGCNN was the best-performing model under transfer learning. Although the point cloud models were competitive in the single-source design, no clear difference was observed between the depth and point cloud approaches in the multi-farm scenarios. Further investigation with more diverse farm environments and additional data modalities is needed to establish more generalizable guidelines for selecting appropriate CV approaches for on-farm BW prediction systems when target data are scarce and limited.

\newpage 
\bibliographystyle{apalike} 
\bibliography{BW_prediction}

@article{yosinski2014transferable,
  title={How transferable are features in deep neural networks?},
  author={Yosinski, Jason and Clune, Jeff and Bengio, Yoshua and Lipson, Hod},
  journal={Advances in Neural Information Processing Systems},
  volume={27},
  year={2014}
}

@inproceedings{lin2014microsoft,
  title={Microsoft COCO: Common objects in context},
  author={Lin, Tsung-Yi and Maire, Michael and Belongie, Serge and Hays, James and Perona, Pietro and Ramanan, Deva and Doll{\'a}r, Piotr and Zitnick, C Lawrence},
  booktitle={European Conference on Computer Vision},
  pages={740--755},
  year={2014},
  organization={Springer}
}

@inproceedings{van2010time,
  title={Time series analysis of live weight as health indicator},
  author={van der Tol, Rik and van der Kamp, Arjen},
  booktitle={First North American Conference on Precision Dairy Management},
  pages={230--231},
  year={2010}
}

@article{qiao2021intelligent,
  title={{Intelligent perception for cattle monitoring: A review for cattle identification, body condition score evaluation, and weight estimation}},
  author={Qiao, Yongliang and Kong, He and Clark, Cameron and Lomax, Sabrina and Su, Daobilige and Eiffert, Stuart and Sukkarieh, Salah},
  journal={Computers and Electronics in Agriculture},
  volume={185},
  pages={106143},
  year={2021},
  publisher={Elsevier}
}

@article{yu2021forecasting,
  title={{Forecasting dynamic body weight of nonrestrained pigs from images using an RGB-D sensor camera}},
  author={Yu, Haipeng and Lee, Kiho and Morota, Gota},
  journal={Translational Animal Science},
  volume={5},
  number={1},
  pages={txab006},
  year={2021},
  publisher={Oxford University Press US}
}

@article{bi2023depth,
  title={{Depth video data-enabled predictions of longitudinal dairy cow body weight using thresholding and Mask R-CNN algorithms}},
  author={Bi, Ye and Campos, Leticia M and Wang, Jin and Yu, Haipeng and Hanigan, Mark D and Morota, Gota},
  journal={Smart Agricultural Technology},
  volume={6},
  pages={100352},
  year={2023},
  publisher={Elsevier}
}

@article{xie2024novel,
  title={{A novel approach based on a modified mask R-CNN for the weight prediction of live pigs}},
  author={Xie, Chuanqi and Cang, Yuji and Lou, Xizhong and Xiao, Hua and Xu, Xing and Li, Xiangjun and Zhou, Weidong},
  journal={Artificial Intelligence in Agriculture},
  volume={12},
  pages={19--28},
  year={2024},
  publisher={Elsevier}
}

@article{hou2023body,
  title={{Body weight estimation of beef cattle with 3D deep learning model: PointNet++}},
  author={Hou, Zixia and Huang, Lyuwen and Zhang, Qi and Miao, Yuanshuang},
  journal={Computers and Electronics in Agriculture},
  volume={213},
  pages={108184},
  year={2023},
  publisher={Elsevier}
}

@article{yang2022automated,
  title={{Automated measurement of dairy cows body size via 3D point cloud data analysis}},
  author={Yang, Guangyuan and Xu, Xingshi and Song, Lei and Zhang, Qianru and Duan, Yuanchao and Song, Huaibo},
  journal={Computers and Electronics in Agriculture},
  volume={200},
  pages={107218},
  year={2022},
  publisher={Elsevier}
}

@inproceedings{liu2022convnet,
  title={A convnet for the 2020s},
  author={Liu, Zhuang and Mao, Hanzi and Wu, Chao-Yuan and Feichtenhofer, Christoph and Darrell, Trevor and Xie, Saining},
  booktitle={Proceedings of the IEEE/CVF conference on computer vision and pattern recognition},
  pages={11976--11986},
  year={2022}
}

@inproceedings{deng2009imagenet,
  title={{Imagenet: A large-scale hierarchical image database}},
  author={Deng, Jia and Dong, Wei and Socher, Richard and Li, Li-Jia and Li, Kai and Fei-Fei, Li},
  booktitle={2009 IEEE conference on computer vision and pattern recognition},
  pages={248--255},
  year={2009},
  organization={IEEE}
}

@article{mehta2021mobilevit,
  title={Mobilevit: light-weight, general-purpose, and mobile-friendly vision transformer},
  author={Mehta, Sachin and Rastegari, Mohammad},
  journal={arXiv preprint arXiv:2110.02178},
  year={2021}
}

@article{wang2019dynamic,
  title={Dynamic graph cnn for learning on point clouds},
  author={Wang, Yue and Sun, Yongbin and Liu, Ziwei and Sarma, Sanjay E and Bronstein, Michael M and Solomon, Justin M},
  journal={ACM Transactions on Graphics (tog)},
  volume={38},
  number={5},
  pages={1--12},
  year={2019},
  publisher={Acm New York, NY, USA}
}

@inproceedings{qi2017pointnet,
  title={{Pointnet: Deep learning on point sets for 3d classification and segmentation}},
  author={Qi, Charles R and Su, Hao and Mo, Kaichun and Guibas, Leonidas J},
  booktitle={Proceedings of the IEEE conference on computer vision and pattern recognition},
  pages={652--660},
  year={2017}
}

@article{qi2017pointnet++,
  title={{PointNet++: Deep hierarchical feature learning on point sets in a metric space}},
  author={Qi, Charles Ruizhongtai and Yi, Li and Su, Hao and Guibas, Leonidas J},
  journal={Advances in Neural Information Processing Systems},
  volume={30},
  year={2017}
}

@article{wang2024shinyanimalcv,
  title={{ShinyAnimalCV: open-source cloud-based web application for object detection, segmentation, and three-dimensional visualization of animals using computer vision}},
  author={Wang, Jin and Hu, Yu and Xiang, Lirong and Morota, Gota and Brooks, Samantha A and Wickens, Carissa L and Miller-Cushon, Emily K and Yu, Haipeng},
  journal={Journal of Animal Science},
  volume={102},
  pages={skad416},
  year={2024},
  publisher={Oxford University Press US}
}

@article{de2025animalmotionviz,
  title={{AnimalMotionViz: An interactive software tool for tracking and visualizing animal motion patterns using computer vision}},
  author={De Castro, Angelo L and Wang, Jin and Bonney-King, Jessica G and Morota, Gota and Miller-Cushon, Emily K and Yu, Haipeng},
  journal={JDS Communications},
  volume={6},
  number={3},
  pages={416--421},
  year={2025},
  publisher={Elsevier}
}

@article{gonccalves2025computer,
  title={{Computer vision analysis of luteal color Doppler ultrasonography for early and automated pregnancy diagnosis in Bos taurus beef cows}},
  author={Gon{\c{c}}alves, Lucas Melo and Fontes, Pedro Levy Piza and Alves, Anderson Antonio Carvalho},
  journal={Journal of Animal Science},
  pages={skaf166},
  year={2025},
  publisher={Oxford University Press US}
}

@article{paudel2023deep,
  title={Deep learning models to predict finishing pig weight using point clouds},
  author={Paudel, Shiva and de Sousa, Rafael Vieira and Sharma, Sudhendu Raj and Brown-Brandl, Tami},
  journal={Animals},
  volume={14},
  number={1},
  pages={31},
  year={2023},
  publisher={MDPI}
}

@article{ruchay2024barn,
  title={On-barn cattle facial recognition using deep transfer learning and data augmentation},
  author={Ruchay, Alexey and Kolpakov, Vladimir and Guo, Hao and Pezzuolo, Andrea},
  journal={Computers and Electronics in Agriculture},
  volume={225},
  pages={109306},
  year={2024},
  publisher={Elsevier}
}

@article{machuve2022poultry,
  title={Poultry diseases diagnostics models using deep learning},
  author={Machuve, Dina and Nwankwo, Ezinne and Mduma, Neema and Mbelwa, Jimmy},
  journal={Frontiers in Artificial Intelligence},
  volume={5},
  pages={733345},
  year={2022},
  publisher={Frontiers Media SA}
}

@article{rodriguez2019estimating,
  title={Estimating body condition score in dairy cows from depth images using convolutional neural networks, transfer learning and model ensembling techniques},
  author={Rodriguez Alvarez, Juan and Arroqui, Mauricio and Mangudo, Pablo and Toloza, Juan and Jatip, Daniel and Rodriguez, Juan M and Teyseyre, Alfredo and Sanz, Carlos and Zunino, Alejandro and Machado, Claudio and others},
  journal={Agronomy},
  volume={9},
  number={2},
  pages={90},
  year={2019},
  publisher={MDPI}
}

@article{bi2025industry,
  title={Industry-scale prediction of video-derived pig body weight using efficient convolutional neural networks and vision transformers},
  author={Bi, Ye and Huang, Yijian and Xuan, Jianhua and Morota, Gota},
  journal={Biosystems Engineering},
  volume={257},
  pages={104243},
  year={2025},
  publisher={Elsevier}
}

@article{miller2019using,
  title={{Using 3D imaging and machine learning to predict liveweight and carcass characteristics of live finishing beef cattle}},
  author={Miller, Gemma A and Hyslop, James J and Barclay, David and Edwards, Andrew and Thomson, William and Duthie, Carol-Anne},
  journal={Frontiers in Sustainable Food Systems},
  volume={3},
  pages={30},
  year={2019},
  publisher={Frontiers Media SA}
}

@article{espejo2020towards,
  title={Towards weeds identification assistance through transfer learning},
  author={Espejo-Garcia, Borja and Mylonas, Nikos and Athanasakos, Loukas and Fountas, Spyros and Vasilakoglou, Ioannis},
  journal={Computers and Electronics in Agriculture},
  volume={171},
  pages={105306},
  year={2020},
  publisher={Elsevier}
}

@inproceedings{gjergji2020deep,
  title={Deep learning techniques for beef cattle body weight prediction},
  author={Gjergji, Mikel and de Moraes Weber, Vanessa and Silva, Luiz Ot{\'a}vio Campos and da Costa Gomes, Rodrigo and De Ara{\'u}jo, Thiago Lu{\'\i}s Alves Campos and Pistori, Hemerson and Alvarez, Marco},
  booktitle={2020 International Joint Conference on Neural Networks (IJCNN)},
  pages={1--8},
  year={2020},
  organization={IEEE}
}

@article{dohmen2021image,
  title={Image-based body mass prediction of heifers using deep neural networks},
  author={Dohmen, Roel and Catal, Cagatay and Liu, Qingzhi},
  journal={Biosystems Engineering},
  volume={204},
  pages={283--293},
  year={2021},
  publisher={Elsevier}
}

@article{wang2025impact,
  title={Impact of cross-validation designs on cattle behavior prediction using machine learning and deep learning models with tri-axial accelerometer data},
  author={Wang, Jin and Yu, Ziwen and Chebel, Ricardo C and Yu, Haipeng},
  journal={Smart Agricultural Technology},
 volume={12},  
pages={101483},
  year={2025},
}

@article{li2023transforming,
  title={Transforming medical imaging with Transformers? A comparative review of key properties, current progresses, and future perspectives},
  author={Li, Jun and Chen, Junyu and Tang, Yucheng and Wang, Ce and Landman, Bennett A and Zhou, S Kevin},
  journal={Medical image analysis},
  volume={85},
  pages={102762},
  year={2023},
  publisher={Elsevier}
}

@inproceedings{li2022improving,
  title={Improving adversarial robustness of 3D point cloud classification models},
  author={Li, Guanlin and Xu, Guowen and Qiu, Han and He, Ruan and Li, Jiwei and Zhang, Tianwei},
  booktitle={European conference on computer vision},
  pages={672--689},
  year={2022},
  organization={Springer}
}

@article{tang2024non,
  title={A Non-Destructive Detection and Grading Method of the Internal Quality of Preserved Eggs Based on an Improved ConvNext},
  author={Tang, Wenquan and Zhang, Hao and Chen, Haoran and Fan, Wei and Wang, Qiaohua},
  journal={Foods},
  volume={13},
  number={6},
  pages={925},
  year={2024},
  publisher={MDPI}
}

@article{ohamouddou2025ms,
  title={MS-DGCNN++: A Multi-Scale Fusion Dynamic Graph Neural Network with Biological Knowledge Integration for LiDAR Tree Species Classification},
  author={Ohamouddou, Said and Afia, Abdellatif El and Afia, Hanaa El and Chiheb, Raddouane},
  journal={arXiv preprint arXiv:2507.12602},
  year={2025}
}

@article{kashiha2014automatic,
  title={Automatic weight estimation of individual pigs using image analysis},
  author={Kashiha, Mohammadamin and Bahr, Claudia and Ott, Sanne and Moons, Christel PH and Niewold, Theo A and {\"O}dberg, Frank O and Berckmans, Daniel},
  journal={Computers and Electronics in Agriculture},
  volume={107},
  pages={38--44},
  year={2014},
  publisher={Elsevier}
}

@article{poursaberi2010real,
  title={Real-time automatic lameness detection based on back posture extraction in dairy cattle: Shape analysis of cow with image processing techniques},
  author={Poursaberi, Ahmad and Bahr, Claudia and Pluk, Arno and Van Nuffel, Annelies and Berckmans, Daniel},
  journal={Computers and electronics in agriculture},
  volume={74},
  number={1},
  pages={110--119},
  year={2010},
  publisher={Elsevier}
}

% \newpage
\section*{Author contribution statement}
\textbf{Jin Wang}: Investigation, Methodology, Software, Formal analysis, Visualization, Writing – original draft, Writing – review \& editing.
\textbf{Angelo De Castro}: Investigation, Methodology, Formal analysis, Writing – review \& editing.
\textbf{Yuxi Zhang}: Investigation, Writing – review \& editing.
\textbf{Lucas Basolli Borsatto}: Investigation, Writing – review \& editing.
\textbf{Yuechen Guo}: Investigation, Writing – review \& editing.
\textbf{Victoria Bastos Primo}: Investigation, Writing – review \& editing.
\textbf{Ana Beatriz Montevecchio Bernardino}: Investigation, Writing – review \& editing.
\textbf{Gota Morota}: Methodology, Writing – review \& editing.
\textbf{Ricardo C. Chebel}: Investigation, Methodology, Writing – review \& editing.
\textbf{Haipeng Yu}: Conceptualization, Investigation, Methodology, Writing – original draft, Writing – review \& editing, Supervision, Project administration, Funding acquisition.

%\section*{Acknowledgments}

\section*{Funding}
This work was supported by the University of Florida startup funds to H.Y. 

\section*{Conflict of interest}
The authors declare that there is no conflict of interest.

\newpage
\section*{Tables}
% Table 1
\begin{table}[H]
\begin{center}
\caption{Camera setup parameters and installation positions at each farm.}
\label{camera_intrinsics}
\scalebox{0.9}{
\begin{tabular}{lccccccc} \hline \hline
\multicolumn{1}{l}{\textbf{Farm}} &
\multicolumn{1}{c}{\textbf{\boldmath$f_x$ (px)}} &
\multicolumn{1}{c}{\textbf{\boldmath$f_y$ (px)}} &
\multicolumn{1}{c}{\textbf{\boldmath$c_x$ (px)}} &
\multicolumn{1}{c}{\textbf{\boldmath$c_y$ (px)}} &
\multicolumn{1}{c}{\textbf{Width (px)}} &
\multicolumn{1}{c}{\textbf{Height (px)}} &
\multicolumn{1}{c}{\textbf{\boldmath$H_{\text{ground}}$ (m)}} \\ \hline
Large-farm   & 388.48 & 388.48 & 326.86 & 240.69 & 640 & 480 & 2.52 \\
Medium-farm     & 386.19 & 385.79 & 326.81 & 246.99 & 640 & 480 & 3.05 \\
Small-farm  & 385.04 & 384.57 & 329.22 & 241.01 & 640 & 480 & 3.00 \\ \hline
\end{tabular}}
\\[0.6em]
\footnotesize{$f_x$, $f_y$: focal lengths in the $x$- and $y$-directions; $c_x$, $c_y$: principal point coordinates; Width, Height: image resolution; $H_{\text{ground}}$: distance from the camera to the ground; px: pixels; and m: meters.}
\end{center}
\end{table}

\newpage
% Table 2
\begin{table}[H]
\begin{center}
\caption{Tuned hyperparameters and their search ranges for both depth image and point cloud models.}
\label{hyperparameters_tuned}
\scalebox{0.95}{
\begin{tabular}{llll}
\toprule
\textbf{Model type} & \textbf{Model} & \textbf{Tuned hyperparameter} & \textbf{Range} \\
\midrule

\multirow{8}{*}{\textbf{Depth image model}}
& \multirow{4}{*}{\textbf{ConvNeXt}}
& Learning rate & \{$10^{-5}$, $10^{-4}$, $10^{-3}$\} \\
& & Dense units & $\{64,\,128,\,192,\,256\}$ \\
& & Dropout & \{$0.1$, $0.2$, $0.3$, $0.4$, $0.5$\} \\
& & Weight decay & \{$10^{-6}$, $10^{-5}$, $10^{-4}$, $10^{-3}$, $10^{-2}$\} \\
\cmidrule(l){2-4}
& \multirow{4}{*}{\textbf{MobileViT}}
& Learning rate & \{$10^{-5}$, $10^{-4}$, $10^{-3}$\} \\
& & Dense units & $\{64,\,128,\,192,\,256\}$ \\
& & Dropout & \{$0.1$, $0.2$, $0.3$, $0.4$, $0.5$\} \\
& & Weight decay & \{$10^{-6}$, $10^{-5}$, $10^{-4}$, $10^{-3}$, $10^{-2}$\} \\
\midrule

\multirow{10}{*}{\textbf{Point cloud model}}
& \multirow{5}{*}{\textbf{DGCNN}}
& Learning rate & \{$10^{-6}$, $10^{-5}$, $10^{-4}$, $10^{-3}$\} \\
& & Dropout & \{$0.2$, $0.3$, $0.4$, $0.5$\} \\
& & $k$ neighbors & $\{15,\,20\}$ \\
& & Weight decay & \{$10^{-5}$, $10^{-4}$, $10^{-3}$, $10^{-2}$\} \\
& & Embedding dims & $\{256,\,512,\,1024\}$ \\
\cmidrule(l){2-4}
& \multirow{5}{*}{\textbf{PointNet}}
& Learning rate & \{$10^{-6}$, $10^{-5}$, $10^{-4}$, $10^{-3}$\} \\
& & Dropout & \{$0.2$, $0.3$, $0.4$, $0.5$\} \\
& & Weight decay & \{$10^{-5}$, $10^{-4}$, $10^{-3}$, $10^{-2}$\} \\
& & Embedding dims & $\{256,\,512,\,1024\}$ \\
& & Feature T-Net & \{on, off\} \\
\bottomrule
\end{tabular}}
\end{center}
\end{table}

\newpage
% Table 3
\begin{table}[H]
\begin{center}
\caption{Body weight prediction performance for the small-farm testing set under single-source, joint learning, and transfer learning strategies. Performance is reported for depth image models (ConvNeXt and MobileViT) and point cloud (PC) models (PointNet and DGCNN). Values are reported as the mean, with the standard error shown in parentheses, across five repeated random subsampling cross-validations. $R^2$ denotes the coefficient of determination, and MAPE denotes the mean absolute percentage error.}
\label{Overall_results}
\scalebox{0.6}{
\begin{tabular}{l l cccccccc}
\toprule
\textbf{Strategy} & 
\textbf{External data scenario} & 
\multicolumn{4}{c}{$\mathbf{R^2}$} & 
\multicolumn{4}{c}{\textbf{MAPE (\%)}} \\
\cmidrule(lr){3-6}\cmidrule(lr){7-10}
 &  & 
\multicolumn{2}{c}{\textbf{Depth image model}} & \multicolumn{2}{c}{\textbf{PC model}} &
\multicolumn{2}{c}{\textbf{Depth image model}} & \multicolumn{2}{c}{\textbf{PC model}} \\
\cmidrule(lr){3-4}\cmidrule(lr){5-6}\cmidrule(lr){7-8}\cmidrule(lr){9-10}
 &  & 
\textbf{ConvNeXt} & \textbf{MobileViT} & \textbf{PointNet} & \textbf{DGCNN} & 
\textbf{ConvNeXt} & \textbf{MobileViT} & \textbf{PointNet} & \textbf{DGCNN} \\
\midrule

\multirow{3}{*}{Single-source} & 
&   &  &   &  &  &  &  &  \\
& No external data   & 0.30 (0.03) & 0.16 (0.04) & 0.37 (0.09) & 0.39 (0.11) & 9.32 (0.93) & 10.02 (0.90) & 8.19 (1.05) & 8.48 (0.99) \\
&   &  &   &  &  &  &  &  \\
\midrule

\multirow{5}{*}{Joint learning} & 
&   &  &   &  &  &  &  &  \\
& Medium-farm & 0.64 (0.06) & 0.37 (0.08) & 0.46 (0.05) & 0.75 (0.05) & 6.78 (1.10) & 8.62 (1.13) & 8.12 (1.03) & 5.63 (0.66) \\
& Large-farm & 0.57 (0.05) & 0.30 (0.10) & 0.68 (0.04) & 0.77 (0.01) & 7.21 (1.05) & 10.42 (0.93) & 7.52 (0.71) & 5.38 (0.56) \\
& Medium + Large farm & 0.65 (0.04) & 0.31 (0.08) & 0.73 (0.03) & 0.74 (0.03) & 6.64 (0.74) & 13.37 (1.11) & 5.74 (0.61) & 5.83 (0.49) \\
&   &  &   &  &  &  &  &  \\
\midrule

\multirow{5}{*}{Transfer learning} & 
&   &  &   &  &  &  &  &  \\
& Medium-farm & 0.46 (0.05) & 0.68 (0.07) & 0.70 (0.01) & 0.70 (0.03) & 7.81 (0.91) & 6.71 (1.20) & 6.89 (0.70) & 5.73 (0.76) \\
& Large-farm & 0.51 (0.01) & 0.78 (0.03) & 0.71 (0.02) & 0.77 (0.02) & 7.60 (0.96) & 5.98 (1.10) & 6.74 (0.73) & 5.07 (0.67) \\
& Medium + Large farm      & 0.67 (0.05) & 0.82 (0.02) & 0.68 (0.04) & 0.79 (0.03) & 6.48 (1.08) & 5.41 (1.01) & 6.79 (0.87) & 5.44 (0.61) \\
&   &  &   &  &  &  &  &  \\
\bottomrule
\end{tabular}}
\end{center}
\end{table}

\newpage
\section*{Figures}
%Fig1

\newpage
%Fig2

\newpage
%Fig3

\end{document}